\documentclass[letterpaper, 10 pt, conference]{ieeeconf}  

\IEEEoverridecommandlockouts                              

\usepackage{graphicx}
\usepackage{algorithm}
\usepackage{tabularx}
\usepackage{amsmath, amssymb}
\usepackage{subcaption}
\usepackage{cite}
\usepackage{booktabs}
\usepackage{algpseudocode}
\usepackage[export]{adjustbox} 
\usepackage{tikz} 
\usepackage{svg}
\usepackage{multirow}

\usepackage{makecell} 
\usepackage[hidelinks]{hyperref}
\let\labelindent\relax
\usepackage{enumitem}

\newcommand{\eqdef}{:=}

\newcommand{\commentout}[1]{}

\title{MR-STORM: Scalable Multi-Arm Control By Distributed MPC}
\author{Dan Evron$^{1}$, 
         Elias~Goldsztejn$^{1}$, 
         Dan R.~Suissa$^{1}$,
         Ronen I. Brafman$^{1}$ 
 \thanks{This research was supported by Ben-Gurion University of the Negev through the Agricultural, Biological and Cognitive Robotics Initiative,
 the Marcus Endowment Fund and the Helmsley Charitable Trust.}
\thanks{$^{1}$
 Foundations of AI Institute, 
 Stein Faculty of Computer and Information Science,
 Ben-Gurion University of the Negev.
        \{evrond,eliasgol,danrouve\}@post.bgu.ac.il, brafman@bgu.ac.il}}%

\usepackage{tikz}
\AddToHook{shipout/firstpage}{%
  \begin{tikzpicture}[remember picture, overlay]
    \node[anchor=north, align=center, text width=17cm, font=\fontfamily{ptm}\fontsize{7}{9}\selectfont] at ([yshift=-1cm]current page.north) {%
      © 2026 IEEE. Personal use of this material is permitted. Permission from IEEE must be obtained for all other uses, in any current or future media, including reprinting/republishing this material for advertising or promotional purposes, creating new collective works, for resale or redistribution to servers or lists, or reuse of any copyrighted component of this work in other works.
    };
  \end{tikzpicture}%
}
\begin{document}
\maketitle

\thispagestyle{empty}
\pagestyle{empty}

\begin{abstract}

Centralized motion planning algorithms struggle to scale in multi-arm manipulation, where inter-arm coupling and tight workspaces amplify complexity.
We introduce MR-STORM, a distributed sampling-based MPC framework that leverages massively parallel GPU sampling for arm control. Our approach integrates inter-arm collision avoidance via plan-sharing and a dynamic priority scheme to maintain task efficiency. We show empirically that MR-STORM outperforms existing baselines in complex simulation environments and that it transfers effectively to physical hardware. Code, demos and data are available at: \url{https://roboworkshop.github.io/multi-robot-mpc/}

\end{abstract}

\section{INTRODUCTION}

\label{introduction:the-problem}
We address motion generation for multiple high-DoF arms in uncertain, dynamic environments. Whether for humanoids or industrial sorting lines, robots must operate effectively in dense, shared workspaces. While safety around static obstacles is manageable through long-range planning, dynamic scenes demand agile, collision-avoidant behavior to navigate inherent unpredictability. This includes avoiding human co-workers, moving items on a conveyor belt, or tracking evolving targets. In all situations, these arms must operate with zero tolerance for error. A primary challenge lies in deploying multiple arms within the same cluttered workspace. The resulting high degree of inter-arm interaction, where the cost of a collision can be fatal, is difficult to manage in static environments and even more so in dynamic settings. This coordination challenge is the focus of our work.

\label{introduction:solution-1-centralized-methods-and-their-drawbacks}


Centralized methods compute joint trajectories for all robots and range from approaches with optimality guarantees~\cite{drrt*, SHARON201540} to more heuristic variants~\cite{PP_87_Lorenzo_p, CBS_MPC}. In both cases, a single coordinator must resolve inter-robot constraints, and the computational load grows quickly with the number of robots.
Moreover, many centralized approaches are global planners that search for complete start-to-goal solutions, often leading to longer and less predictable runtimes than receding-horizon methods. This limits real-time responsiveness in dynamic environments.

Distributed MPC~\cite{christofides2013dmpc,
scattolini2009architectures,luis2020online} provides the natural alternative to centralized joint-space control: each robot solves a local receding-horizon problem while exchanging predicted motions or constraints with neighboring robots. Existing distributed MPC methods have shown that this decomposition can coordinate multiple manipulators in shared workspaces. However, in dense high-DoF arm settings, \textit{optimization-based} DMPC must repeatedly solve nonconvex constrained problems with many pairwise link-level collision constraints, and often requires additional mechanisms to resolve deadlocks or priority conflicts~\cite{gafur2022dynamic,
chen2024deadlock}.
This creates a tension between the formal constraint-handling strengths of DMPC and the low-latency reactive behavior needed when several arms operate in a tightly coupled dynamic workspace. 

\label{introduction:solution-2-STORM}
To address these challenges, we seek a decentralized, reactive algorithm capable of adapting to sudden hazards or goal changes, focusing on \textit{sampling-based} DMPC. 
We build on STORM \cite{Storm}, a state-of-the-art single-manipulator algorithm that leverages GPUs to efficiently implement a parallel, sampling-based, model predictive control (MPC) method \cite{mppi_first_2017}. 
 \label{introduction:drawbacks-of-STORM}
A central limitation of STORM is that it does not consider the predicted trajectories of objects within its MPC model. While this suffices for static or slow-moving obstacles, it fails in the presence of rapidly moving ones, and in particular, additional robotic arms. Unsurprisingly, this issue is exacerbated when the workspace is small and every motion matters.
Our empirical evaluations confirm that this approach indeed leads to high collision  frequency.


\label{introduction:solution3-MR-STORM.Part1-safety}
We overcome this problem by leveraging information generated by STORM during execution:
its control distribution for the next $H$ steps,
which we use to 
model predicted occupancy and account for the dynamic constraints. Specifically, arms communicate their near-term trajectories—derived from the means of their control distributions and incorporate this information in their respective cost functions. We demonstrate that this approach significantly improves safety.

\label{introduction:Solution3-MR-STORM.Part2-task(priority)}
Unlike global planners, STORM is an MPC-based approach and therefore inherently local. Consequently, it is susceptible to local minima, which manifests as deadlocks (complete cessation of movement) or livelocks (persistent oscillations). These states arise when conflicts between goal-reaching and collision-avoidance objectives remain unresolved. We address this issue using a distance-based priority scheme that resolves conflicts dynamically through shared rules, ensuring all computations remain decentralized.

\label{introduction:evaluation}

\label{introduction:main_contribution}
Our main contributions are summarized as follows:
\begin{itemize}[nosep,leftmargin=*]
\item \textbf{MR-STORM:} 
A sampling-based distributed MPC algorithm
for multi-arm manipulation, extending STORM/MPPI\cite{Storm} 
to decentralized coordination in cluttered, dynamic shared workspaces. Our method scales efficiently while maintaining high-frequency control.
\item \textbf{Decentralized Prioritization:} A novel coordination scheme that resolves conflicts via shared rules, ensuring smooth, collision-free motion without sudden stops or a central coordinator.
\item \textbf{Experimental Validation:} Extensive benchmarking across 120 simulated scenarios and on physical hardware, demonstrating superior performance over established baselines using realistic robot models.
\end{itemize}

\label{related-work}
\section{Related Work}

\label{related-work:single-robot-motion-generation}
\paragraph{Single-arm Motion Generation}

\label{related-work:single-arm-motion-generation:problem-definition}
Given an initial configuration $\Theta_{\text{start}} \in \mathcal{C}$ (the configuration space) and either a goal pose (position and quaternion) $g = [g_p, g_q] \in \mathbb{R}^3 \times \mathbb{S}^3$   
indicating the desired state of the end-effector link frame, or a goal configuration $\Theta_{\text{end}}$,  motion-generation algorithms seek a collision-free trajectory within $\mathcal{C}$ that satisfies kinematic and dynamic constraints while minimizing task-specific costs.
In this work, we focus on the former case, where the goal pose of the end-effector link is specified. 

\label{related-work:single-arm-motion-generation:method-types}

\label{related-work:single-arm-motion-generation:classical(graph-based)-methods}
\textit{Graph-based planning} methods construct a discrete connectivity structure (e.g., uniform grids, visibility or Voronoi graphs, or probabilistic roadmaps~\cite{PRM}) and apply search algorithms such as A*.
\textit{Trajectory optimization} methods \cite{ratliff_chomp_2009,stomp_2011,Curobo} operate 
in the continuous space of parameterized state trajectories and minimize cost functionals subject to dynamics constraints. CHOMP \cite{ratliff_chomp_2009} uses gradient-based optimization with signed distance fields for collision checking. STOMP \cite{stomp_2011} uses stochastic perturbations to handle non-differentiable costs. cuRobo \cite{Curobo} enhances STOMP, leveraging GPU-accelerated massive parallelism and continuous collision checking.
These methods optimize
complete motions offline, 
suffering from latency and open-loop brittleness, 
and are 
less suitable for dynamic environments or model uncertainties compared to reactive, receding-horizon paradigms.



\label{related-work:single-arm-motion-generation:mpc}
\textit{Model Predictive Control (MPC)} formulates motion generation as a finite-horizon optimal control problem solved repeatedly in a receding-horizon manner. 
At each control step, MPC optimizes a control sequence $\mathbf{u}$
using an explicit system dynamics model 
$
x_{t+1} = f(x_t, u_t),
$
minimizing a cumulative cost over the horizon. Only the first control $u_1$ is executed before the problem is re-solved at the next timestep, enabling continual adaptation to changes in goals or environmental conditions compared to \cite{ratliff_chomp_2009,stomp_2011,Curobo}.
%

\label{related-work:single-arm-motion-generation:mppi}
Model Predictive Path Integral (MPPI) (see~\ref{background:storm:mppi-paradigm}) and related information-theoretic MPC methods replace deterministic optimization with sampling-based updates. 
MPPI performs derivative-free stochastic optimization and relies on parallel rollout evaluation rather than iterative constraint-based optimization. 
%
Building on MPPI~\cite{mppi_first_2017}, STORM~\cite{Storm} is a high-performance framework for robotic arms that leverages GPU-accelerated batched rollouts to optimize for collision avoidance, smoothness, and task-space objectives under joint and velocity constraints. As a low-latency, platform-agnostic optimizer, it is a state-of-the-art solution for reactive high-DoF motion generation,
focusing on single-arm or coupled planning. MR-STORM extends this method to decentralized multi-arm coordination.

\label{related-work:single-arm-motion-generation:computational_tradeoffs_between_approaches}


\paragraph{Multi-robot motion generation}

\label{related-work:multi-robot-motion-generation:centralized-VS-decentralzied}
Let $\mathcal{C}^{(i)}$ be the configuration space of arm $i$. We distinguish between:
\begin{itemize}[nosep,leftmargin=*]
    \item \textit{Centralized:} A central node computes a joint solution in $\mathcal{C} = \prod \mathcal{C}^{(i)}$. Whether planning is coupled (searching full joint space) or decoupled (searching individual spaces, separately), a single authority resolves conflicts and returns a valid solution within the composite space $\mathcal{C}$.
    \item \textit{Decentralized:} Each arm plans in $\mathcal{C}^{(i)}$ and executes independently. Interactions may be managed via local coordination, but the central node is bypassed entirely.
\end{itemize}
\label{related-work:multi-robot-motion-generation}
Classical search methods can be lifted into the composite configuration space to provide completeness and optimality at the cost of greatly increased configuration space dimension. Therefore, algorithms like \textit{dRRT*} \cite{drrt*} use implicit product roadmaps for efficiency. Yet their performance still degrades as DoF and inter-robot constraints increase. 



\textit{Distributed MPC} (DMPC) extends receding-horizon control to multi-robot systems by decomposing the 
joint problem into local subproblems. Agents exchange predicted trajectories, constraints, or coupling  information~\cite{christofides2013dmpc,scattolini2009architectures,luis2020online}. For multi-manipulator coordination, Gafur et al.~\cite{gafur2022dynamic} formulate distributed nonlinear MPC in which each arm optimizes its own joint-space motion
while accounting for predicted motions of neighboring arms through dynamic collision constraints and deadlock-resolution logic. We follow the same distributed predictive-control principle, with a \textit{sampling-based }DMPC method that uses MPPI's derivative-free stochastic optimization technique, relying on parallel rollout evaluation rather than iterative constraint-based optimization. 

\label{related-work:multi-robot-motion-generation:PP}
\textit{Prioritized planning}~\cite{PP_87_Lorenzo_p, PP_SILVER05_Cooperative_PF} 
sequentially plans for each robot. Higher-priority trajectories are treated as moving obstacles for subsequent agents. Hence, it does not benefit from true parallelism.
Static orderings are sensitive to the chosen sequence and may fail in dense scenarios. Dynamic variants attempt to resolve these weaknesses by re-evaluating or swapping priorities during contention to improve reactivity.

\label{related-work:multi-robot-motion-generation:CBS}
\textit{Conflict-Based Search (CBS)}~\cite{SHARON201540} resolves overlaps via a high-level constraint tree; hybrid variants like CBS-MP~\cite{CBS_MP} and CBS-MPC~\cite{CBS_MPC} extend this to continuous spaces using low-level sampling or MPC. While effective for sparse interactions, the constraint set grows rapidly as trajectories densify, approaching coupled planning complexity. In contrast, our approach avoids this combinatorial explosion by managing interactions via local, priority-aware cost functions, ensuring planning latency remains independent of global conflict counts, even in dense configurations.

\label{related-work:multi-robot-motion-generation:VO}

\textit{Velocity Obstacles} (e.g., ORCA~\cite{ORCA}) ensure safety by selecting velocities outside regions predicted to cause collisions. While highly reactive, their reliance on reciprocal responsibility—the assumption that all agents share the burden of avoidance—often fails in dense settings. In tightly coupled workspaces, these methods are prone to deadlocks—reaching a zero-velocity consensus—or livelocks, where symmetric avoidance logic induces oscillations.
%
%
%
To address this issue, MR-STORM integrates dynamic prioritization into STORM, breaking the symmetry inherent in reciprocal avoidance. 

\section{Background: The STORM Algorithm}
\subsection{MPPI for Arm-Control} 

\label{background:storm:mppi-paradigm}
STORM \cite{Storm} implements Model Predictive Path Integral (MPPI) \cite{mppi_first_2017} for manipulators, supporting non-differentiable costs and joint limits. It achieves low-latency control by evaluating $N$ rollouts in parallel using an approximate dynamics model $f(x,u)$ to sample and select the best trajectory. Each step, STORM repeats the "Sense-Optimize-Act" paradigm:

\label{background:storm:mppi-paradigm:sense}
\textit{\textbf{Sense}}: STORM queries the current joint state $\Theta = [\theta, \dot{\theta}, \ddot{\theta}] \in \mathbb{R}^{d \times 3}$ and obtains an approximate representation of obstacles $\hat{\mathcal{O}} \in \mathbb{R}^{d_O}$. These, along with the goal pose $g$, form the vectorized state $x \in \mathbb{R}^{3d + d_O + 7}$.

\label{background:storm:mppi-paradigm:optimize}
\textit{\textbf{Optimize}}: STORM maintains a multivariate Gaussian control distribution $\pi_\phi \eqdef \prod_{h=1}^{H}\pi_{\phi,h}$ over an $H$-step horizon. Parameters $\phi \eqdef [[\mu_{1},...,\mu_{H}],[\Sigma_{1},...,\Sigma_{H}]]$ represent means and covariances $\phi_h\eqdef [\mu_h\in \mathbb{R}^{d}, \Sigma_h\in \mathbb{R}^{d\times d}]$ for each step $h$, where $d$ is the number of actuated DoFs. Sampling from $\pi_\phi$ is equivalent to drawing $H$ consecutive, independent $d$-dimensional Gaussian samples.

The optimization cycle repeats $K$ times. MPPI draws $N$ control sequences $\mathbf{U} \in \mathbb{R}^{N \times H \times d}$, where $\mathbf{U}_{n,h} = \mu_{n,h} + \varepsilon_{n,h}$ and $\varepsilon_{n,h} \sim \mathcal{N}(0, \Sigma_h)$. In the STORM framework, these commands $\mathbf{U}_{n,h}$ represent desired joint accelerations. Exploiting the mutual independence of these sequences, STORM parallelizes this sampling process on a GPU.

The MPPI layer processes each control sequence $\mathbf{U}_n = [\mathbf{U}_{n,h}]_{h=1}^{H} \in \mathbb{R}^{H \times d}$ through a dynamics model $f$ to obtain $[\mathbf{\hat C}_n, \mathbf{\hat X}_n] = f(\mathbf{U}_n)$. The resulting "rollout" $\mathbf{\hat X}_n \in \mathbb{R}^{H\times d \times 3}$, approximates the $H$-step joint-state trajectory, where $\mathbf{\hat X}_{n,h} = [\hat{\theta}_{n,h}, \hat{\dot \theta}_{n,h}, \hat{\ddot{\theta}}_{n,h}] \in \mathbb{R}^{d \times 3}$ contains estimated positions, velocities, and accelerations. $\mathbf{\hat C}_n \in \mathbb{R}^H$ represents the corresponding stage costs (see \ref{cost_function}). This 
is parallelized across $N$ sequences on GPU, exploiting   rollout independence. 

Next, $\pi_\phi$ is updated by computing the total discounted rollout cost $\hat C_n \in \mathbb{R}$:
\begin{equation}
\hat C_n = \sum_{h=1}^{H-1} \gamma^{h-1}\mathbf{\hat C}_{n,h} + \gamma^{H-1}\hat q_{n,H},
\end{equation}
where $\mathbf{\hat C}_{n,h}$ represents stage costs and $\hat q_{n,H}$ is the terminal-cost approximation for the "cost-to-go" beyond horizon $H$ (See~\cite{Storm}.2). 
Note that, in general, $\mathbf{\hat C}_n \neq \hat C_n$.

A weight $w_n$ is assigned to each trajectory, where lower costs yield higher weights, 
so that better performing control sequences 
exert greater influence on the updated distribution.

\begin{equation}
\label{eq:storm_weights_single}
w_n \;=\; \frac{\exp\!\big(-\hat C_n/\lambda\big)}{\sum_{m=1}^{N}\exp\!\big(-\hat C_m/\lambda\big)}.
\end{equation}
$\lambda$ is the temperature hyper-parameter.
Next, we compute
\begin{equation}
\label{eq:storm_weighted_stats_single}
\tilde\mu_{h}=\sum_{n=1}^{N} w_{n}\,\mathbf{U}_{n,h}\ \ 
\tilde\Sigma_{h}=\sum_{n=1}^{N} w_n\big(\mathbf{U}_{n,h}-\tilde\mu_{h}\big)\big(\mathbf{U}_{n,h}-\tilde\mu_{h}\big)^\top
\end{equation}
Finally, parameters are updated via exponential moving averages with step sizes $\alpha_\mu, \alpha_\Sigma \in (0,1]$:
\begin{equation}
\label{eq:update-step}
\begin{aligned}
\mu_{h} &\leftarrow (1-\alpha_\mu)\,\mu_{h} + \alpha_\mu\,\tilde\mu_{h}, \\
\Sigma_{h} &\leftarrow (1-\alpha_\Sigma)\,\Sigma_{h} + \alpha_\Sigma\,\tilde\Sigma_{h}.
\end{aligned}
\end{equation}

\label{storm_paradigm_act}
\textit{\textbf{Act}}: 
STORM applies the command $\mathbf{U}_{*} \eqdef \mathbf{U}[n^{*},0]$ from the first step of the minimum-cost trajectory $n^{*} \eqdef \arg\min_{n} \mathbf{\hat C}_n$ at the final optimization step ($k=K$).

\label{summary-and-shifting-motivation}
This paradigm repeats throughout robot operation. At the start of each optimization cycle ($k=1$), we apply \textit{left-shifting}: $\phi_{h}\gets\phi_{h+1} \ \forall h\leq H-1$. This operation discards the previously executed step's distribution; see \cite{Storm}.2 for details.

\subsection{Cost function} 
\label{cost_function}
STORM evaluates $\mathbf{\hat C}$ using an arm-specific cost function that balances task success, safety, and kinematic health:
\begin{align} \label{storm_cost_with_weights}
\mathbf{\hat C}_{n,h} &= w_p \underbrace{\mathbf{\hat C}_{\mathrm{pose}}(\hat{ee}_{n,h}, g)}_{\text{goal reaching}}  
+ w_s \underbrace{\mathbf{\hat C}_{\mathrm{stop}}(\hat{\dot \theta}_{n,h})}_{\text{safe stopping}} \notag \\
&\quad + w_j \underbrace{\mathbf{\hat C}_{\mathrm{joint}}(\hat{ \theta}_{n,h})}_{\text{joint limits}}  
+ w_m \underbrace{\mathbf{\hat C}_{\mathrm{manip}}(\hat{\theta}_{n,h})}_{\text{manipulability}} \notag \\
&\quad + w_c (\underbrace{\mathbf{\hat C}_{\mathrm{coll}}(\hat{ \theta}_{n,h},\hat {\mathcal{O}})+\mathbf{\hat C}_{\text{self-coll}}(\hat{ \theta}_{n,h})}_{\text{collision avoidance}})
\end{align}
%
%
%
%
$\mathbf{\hat C}_{\mathrm{pose}}$ captures end-effector distance/orientation error to goal $g$; $\mathbf{\hat C}_{\mathrm{stop}}$: velocities exceeding safe stopping limits; $\mathbf{\hat C}_{\mathrm{joint}}$: joint soft-limits; $\mathbf{\hat C}_{\mathrm{manip}}$: singularity avoidance via Jacobian ellipsoid volume; and $\mathbf{\hat C}_{\mathrm{coll}}, \mathbf{\hat C}_{\text{self-coll}}$:  penalize for environment and self collisions respectively. A non-negative weight $w_*$ is associated with each of the functions.

\label{mr-storm}
\section{Distributing STORM for Multi-Arm Control}
MR-STORM extends STORM to multi-arm systems by adding a coordination cost based on neighbor trajectories and a dynamic prioritization scheme. This architecture enhances safety with negligible computational overhead. It is also fully decentralized; each arm independently executes STORM with our modifications. 
Our priority-aware scheme ensures scalable coordination without the exponential complexity of centralized approaches.

\label{mr:storm-plan-communication}

\label{mr:storm-plan-communication}
\subsection{Avoiding Inter-Arm Collisions with Plan Communication} 

STORM treats obstacles as static over its $H$-step horizon, updating poses only between MPC steps without predicting their dynamics in $f$ and employs sphere-tiling~\cite{Curobo} (Fig.~3) for fast parallel collision checking during rollouts. Each arm defines a matrix $S \in \mathbb{R}^{N_S \times 4}$ representing $N_S$ spheres, where rows are vectors of position $(x, y, z)$ and radius $r$ at constant transforms from the arm's links.

While predicting environmental obstacle motion is out of scope (see \ref{Section: limitations}), we leverage arms' maintained control distributions for predictive collision checking, dramatically improving safety with minimal computational effort.
This is done as follows: 
After arm $i$ updates its control distribution $\phi^{(i)}$, we apply $f^{(i)}$ to the distribution's means $\mu^{(i)}$ to obtain $H$ joint positions. Through forward kinematics, we then derive the poses of the robot's link frames and obtain their corresponding collision spheres $S^{(i)} \in \mathbb{R}^{N_S^{(i)} \times 4}$ by applying constant transforms with respect to the links. This sequence yields the \textit{anticipated plan} $\hat P_\phi^{(i)} \in \mathbb{R}^{N_S^{(i)} \times 4 \times H}$, which is published to other arms via ``pub-sub'' (publisher-subscriber) mechanisms, utilizing either low-latency shared memory or standard message-passing topics. Finally, other arms incorporate $\hat P_\phi^{(i)}$ into their models by adding a cost term penalizing collisions with these estimated occupancies.

\label{new_cost_fn}
Let $\rho_b(b_1, b_2)$ denote the minimum Euclidean distance ($l_2$ norm) between two general geometric bodies. We define $\hat S_{n,h,l}^{(j)} \in \mathbb{R}^4$ as the position and radius of the $l$-th collision sphere for arm $j$ at step $h$ of rollout $n$, and $(\hat P_{\phi})_{h,k}^{(i)} \in \mathbb{R}^4$ as the $k$-th sphere in the anticipated plan of robot $i$ at the same horizon step $h$. The distance between rollout $n$ of robot $j$ and the plan of robot $i$ at step $h$ is defined as:

\begin{equation}
\rho_b\big(\hat S^{(j)}_{n,h}, (\hat P_{\phi})_h^{(i)}\big) \eqdef \min_{l,k} \rho_b\big(\hat S^{(j)}_{n,h,l}, (\hat P_{\phi})_{h,k}^{(i)}\big)
\end{equation}
That is, the minimal distance between any two spheres
in the descriptions of the arms $j$ and $i$.

MR-STORM augments the original cost function for arm $j$ with a dynamic coordination term:

\begin{equation}
\label{eq:new_cost}
\mathbf{\hat C}_{\text{new}, n,h}^{(j)} = \mathbf{\hat C}_{n,h}^{(j)} + w_d \mathbf{\hat C}_{\text{dyn}, n,h}^{(j)},
\end{equation}
where $\mathbf{\hat C}_{n,h}^{(j)}$ is the original weighted STORM cost from \eqref{storm_cost_with_weights}. 
Define $\mathbf{\hat C}_{\text{dyn}, n,h}^{(j,i)} \eqdef \max\left(0, 1 - \frac{\rho_b(\hat S_{n,h}^{(j)}, \hat P_{\phi, h}^{(i)})}{B}\right)$.
The coordination cost is defined as:

\begin{equation}
\mathbf{\hat C}_{\text{dyn}, n,h}^{(j)} \eqdef \sum_{i \neq j} \mathbf{\hat C}_{\text{dyn}, n,h}^{(j,i)} 
\end{equation}
The safety buffer $B$ defines a proximity threshold for penalties: if $\rho_b(\hat S_{n,h}^{(j)}, \hat P_{\phi, h}^{(i)}) \ge B$ for all $i$, $\mathbf{\hat C}_{\text{dyn}}$ is $0$ and the rollout is not penalized. As the distance decreases below $B$, specifically as $\rho_b \to 0$, the penalty $(1 - \frac{\rho_b}{B}) \to 1$ increases linearly. This maximum penalty at vanishing distance ensures robot $j$ respects its neighbors' anticipated plans.

\textbf{Complexity:} The total number of distance queries for arm $j$ per rollout step is $N_{Q,n,h}^{(j)} = N_S^{(j)} \sum_{i \neq j} N_S^{(i)}$. Over the full horizon and all rollouts, the total calls to $\rho_b$ are $N_Q^{(j)} = H^{(j)} N^{(j)}  N_{Q,n,h}^{(j)}$. Since distance queries between spheres are independent, we leverage GPU parallelization to execute them concurrently. Given that $\rho_b$ for a sphere pair is $\mathcal{O}(1)$, the parallelized runtime complexity remains $\mathcal{O}(1)$, ensuring the coordination cost adds negligible overhead.

\subsection{From Collision Avoidance Only to Task Attainment}
\label{mr-storm:prioritization} 
\begin{figure*}[t]
    \centering
    \includegraphics[width=\textwidth, trim=0 0 0 0, clip]{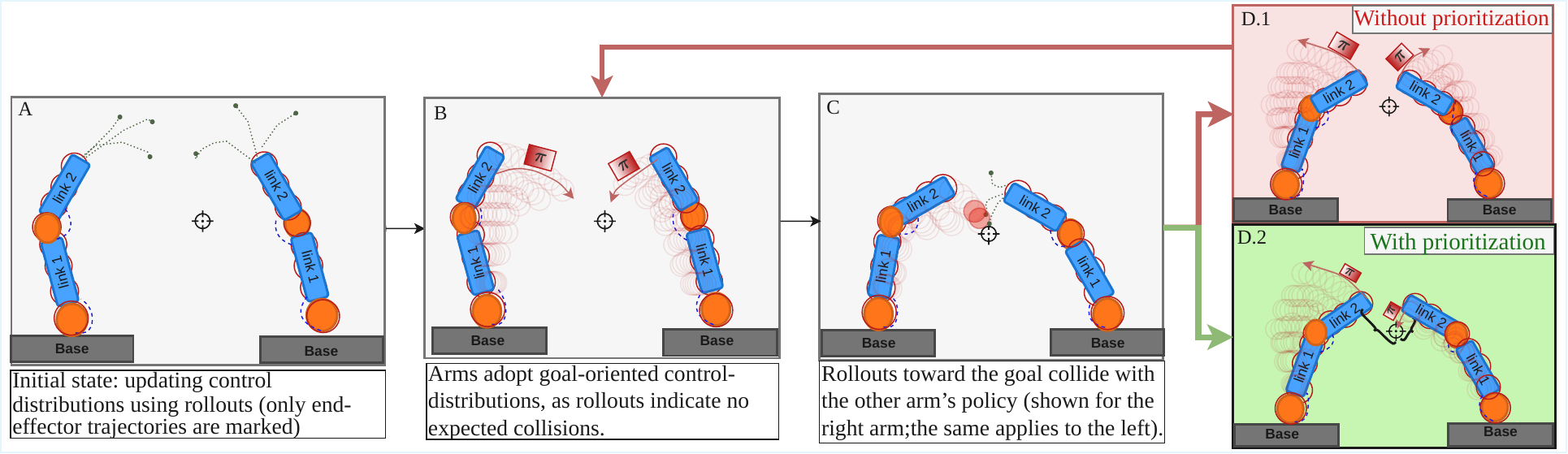}
    \vspace{-15pt}
    \caption{\textbf{MR-STORM priority scheme.} Red opaque spheres indicate current collision occupancy; semi-transparent (pink) spheres show the anticipated plan $\hat P_{\phi}$. Red arrows denote the mean end-effector direction under $\pi_{\phi}$, and green lines show $N=3$ rollout trajectories. \textbf{(A-B)} Both arms use goal-directed control. \textbf{(C)} Overlapping anticipated trajectories triggers arm-to-arm collision cost. \textbf{(D.1) Without prioritization:} Both arms retreat to avoid contact, causing a \textit{livelock}. \textbf{(D.2) With prioritization:} The right arm, closer to its goal, proceeds while the left arm retreats.}
    \label{fig:priority_scheme_example}
    \vspace{-5pt}
\end{figure*}

Consider two arms reaching for the same goal from opposite directions. Initially, the total cost is dominated by goal-reaching terms. However, once anticipated trajectories intersect, the safety-weighted collision cost overrides task completion. Without prioritization, both arms withdraw simultaneously, move forward once the cost vanishes, and repeat the process in a cycle we refer to as \textbf{livelock}. 

To resolve such conflicts, we introduce a multiplicative factor $\alpha^{(j,i)}$ to the dynamic cost:
\begin{equation}
\label{eq:new_cost}
\mathbf{\hat C}_{\text{new}, n,h}^{(j)} = \mathbf{\hat C}_{n,h}^{(j)} + w_d \sum_{i \neq j} \alpha^{(j,i)} \mathbf{\hat C}_{\text{dyn}, n,h}^{(j,i)}
\end{equation}
where the prioritization weight is:
\begin{equation}
\alpha^{(j,i)} \eqdef \left(\frac{\|(ee_p^{(j)} - g_p^{(j)})\|_2}{\|(ee_p^{(i)} - g_p^{(i)})\|_2}\right)^\tau.
\end{equation}
$ \|p_a - p_b\|_2$ denotes the Euclidean distance between points in $\mathbb{R}^3$, while $ee_p$ and $g_p$ represent end-effector and goal positions. This rule prioritizes the robot closer to its target, as it can clear the shared workspace more quickly. Consequently, arms must communicate both their anticipated plans $\hat P_{\phi}$ and their current distances-to-goal.

The exponent $\tau$ acts as a \textbf{trust parameter}. When $\tau = 0$, prioritization is disabled ($\alpha^{(j,i)}= \alpha^{(i,j)}=1 $). For $\tau > 0$, differences in goal proximity are amplified: the "aggressive" prioritized arm ignores its neighbor's plan, while the "cautious" arm assumes the burden of avoidance. This \textbf{smooth prioritization} prevents the sharp cost fluctuations that can destabilize MPPI updates, allowing for a predictable resolution that naturally scales to $A > 2$ arms.

Figure~\ref{fig:priority_scheme_example} describes a typical livelock scenario, comparing our priority-based solution against the baseline without prioritization following the motivating example described above.

\subsection{The MR-STORM Algorithm}

The MR-STORM control loop executed by each robot is detailed in Algorithm~\ref{alg:MRSTORM}. For each agent $i \in \{1, \dots, A\}$, parameters such as $\Theta^{(i)}$, $N^{(i)}$, $K^{(i)}$, $\pi_\phi^{(i)}$, and $S^{(i)}$ retain the meanings defined previously, now localized to the $i^{th}$ robot. We assume a uniform horizon $H^{(i)}=H$ for simplicity. Note that, agents can generally use $min\{H^{(j)}, H^{(i)}\}$ from any received plan. We define $\bar{i}$ as the set of indices $\{1, \dots, A\} \setminus \{i\}$ and $N_S^{(\bar{i})}$ as the total number of collision spheres across all robots in $\bar{i}$. The term $\hat P_{\phi} ^{(\bar i)} \in \mathbb{R}^{N_S^{(\bar i)}\times 4 \times H}$ represents the union of anticipated plans received from all $j \in \bar{i}$. Controls $\phi^{(i)}$ are initialized randomly and arms ignore neighbors whose plans have not yet been published. Given that agents start simultaneously and the algorithm operates at a high frequency, valid $\hat P_{\phi}$ values are assumed 
to arrive with latency small relative to the MPC cycle time.
\begin{figure*}[t]
    \centering
    \includegraphics[width=\linewidth, trim=0 0 0 0, clip]{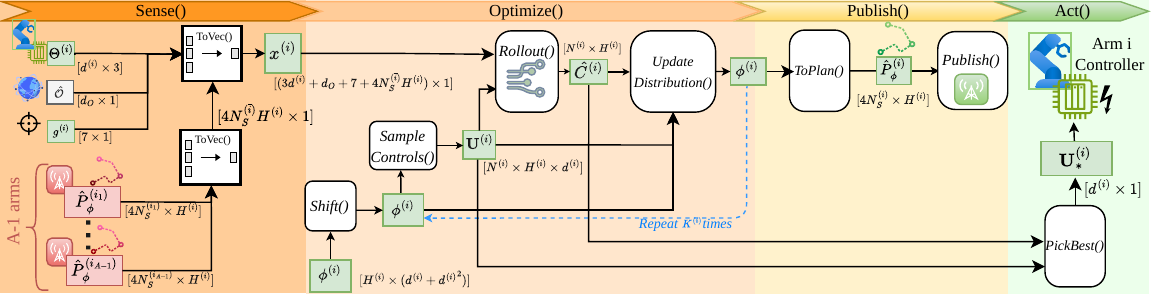}
    \vspace{-15pt}
    \caption{\textbf{MR-STORM control loop }for\textbf{ }arm\textbf{ $i$} (Alg.~\ref{alg:MRSTORM}). \textit{ToVec()} flattens input tensors into a unified vector, while $[A \times B \times \dots]$ denotes a tensor of the specified dimensions. All other components correspond to the steps detailed in Algorithm~\ref{alg:MRSTORM}.}
    \label{fig:mr-storm}
    \vspace{-10pt}
\end{figure*}

\begin{algorithm}[h]
\caption{MR-STORM For Arm  $i$ }\label{alg:MRSTORM}
\begin{algorithmic}[1]
\Statex \textbf{Parameters:} $H^{(i)}, N^{(i)}, K^{(i)}$
\While{True} 
    \State $x^{(i)} \gets {Sense(\hat{\mathcal{O}},\Theta^{(i)}, \hat P_{\phi} ^{(\bar i)})}$ \Comment{Sense} \label{line-get-state}
     \State $\phi^{(i)} \gets Shift(\phi^{(i)})$ \label{line-shift}
    \For{$k = 1 \dots K^{(i)}$} \Comment{Optimize} \label{line-optim-loop}
        \State $\mathbf{ U}^{(i)}\gets {SampleControls}({\phi}^{(i)}, H^{(i)}, N^{(i)})$ \label{line-sample-controls}
        \State $[\mathbf{\hat C_{new}}^{(i)}, \mathbf{\hat X}^{(i)}] \gets {Rollout(\mathbf{U}^{(i)})}$\label{line-rollout-controls}
        \State $\phi^{(i)} \gets {UpdateDistribution}(\mathbf{\hat C_{new}}^{(i)}, \mathbf{\hat X}^{(i)},\mathbf U^{(i)})$\label{line-update-control-distribution}
    \EndFor
    \State $\hat P_\phi^{(i)}\gets ToPlan($$\phi^{(i)})$ \Comment{Publish} \label{line-joints_to_plan}
    \State Publish($\hat P_\phi^{(i)}$)   \label{line-publish-plan}
    \State $\mathbf{U}_{*}^{(i)} \gets {PickBest(\mathbf U^{(i)}})$ \Comment{Act} \label{line-pick-best-rollot-first-action}
    \State {Execute}($\mathbf{U_*^{(i)}}$) \label{line-execute}
\EndWhile
\end{algorithmic}
\end{algorithm}
Each iteration begins with a \textit{sense} phase (line~\ref{line-get-state}) where the state of arm $i$ and the environment are evaluated (\ref{background:storm:mppi-paradigm:sense}). Unlike the original STORM algorithm, the arm now also incorporates the most recent plans from all other robots ($\hat P_{\phi} ^{(\bar i)}$) into its observed state.
Next, the distribution is shifted to align sampling with the present (line~\ref{line-shift}), as per~\ref{summary-and-shifting-motivation}.

The \textit{optimize} phase follows \ref{background:storm:mppi-paradigm:optimize}. As in the original STORM algorithm, controls are sampled (line~\ref{line-sample-controls}), rollouts are generated (line~\ref{line-rollout-controls}), and the distribution is updated (line~\ref{line-update-control-distribution}). However, this iteration employs the cost function $\mathbf{\hat C_{new}}^{(i)}$ from \ref{new_cost_fn}, which incorporates neighbor plans and applies prioritization. After $K^{(i)}$ optimization steps, the agent enters the \textit{publish} phase. In line~\ref{line-joints_to_plan}, the arm computes its anticipated occupancy through forward kinematics and broadcasts the plan to other robots (line~\ref{line-publish-plan}), as described in \ref{mr:storm-plan-communication}. Finally, it proceeds to the \textit{act} phase, selecting the optimal action as per the STORM paradigm (\ref{storm_paradigm_act}). Fig.~\ref{fig:mr-storm} provides a schematic overview to clarify the algorithm's execution.


\section{Empirical Evaluation}

\label{simulated_experiments}
\subsection{Simulated Experiments}

\label{simulated_experiments:setup}
\subsubsection{Setup}

\label{simulated_experiments:setup:tasks}
We compare MR-STORM with several methods 
on goal achievement and collision avoidance. Experiments were conducted in NVIDIA IsaacSim V4.5 on a Linux (Rocky) system with an 8-core AMD® EPYC™ 7302 CPU, 48 GB RAM, and an NVIDIA® RTX™ 4090 GPU. Similar results were achieved on more modest hardware (32 GB RAM, RTX™ 4060). 
Each arm's control rate was 30\,Hz, dictated by our hardware's ability to support the simulation environment. The results and videos show that this is sufficiently fast for good coordination and smooth movement (in the physical experiments, too). However, we verified that the algorithm can work much faster, at 70\,Hz.

Algorithms were tested on 4 multi-arm tasks: \textit{goal reaching (easy/hard)}, \textit{goal following}, and \textit{bin-loading}. Each category included 30 environments across 5 levels of increasing workspace density (inter-arm proximity) to challenge safety. For fairness, all methods were evaluated on identical environments and randomized goal sets, with 1 trial per scenario.

\label{simulated_experiments:setup:robot_spec}
We utilized \(A = 4\) UR5e™ arms (85 cm reach) equipped with magnetic end-effectors to focus specifically on arm motion rather than grasping. Each robot has \(d = 6\) degrees of freedom, totaling 24 for the full system model.

\begin{figure*}[t] 
    \centering
    \begin{subfigure}{0.24\textwidth}
        \includegraphics[width=\linewidth, trim=350 280 350 80, clip]{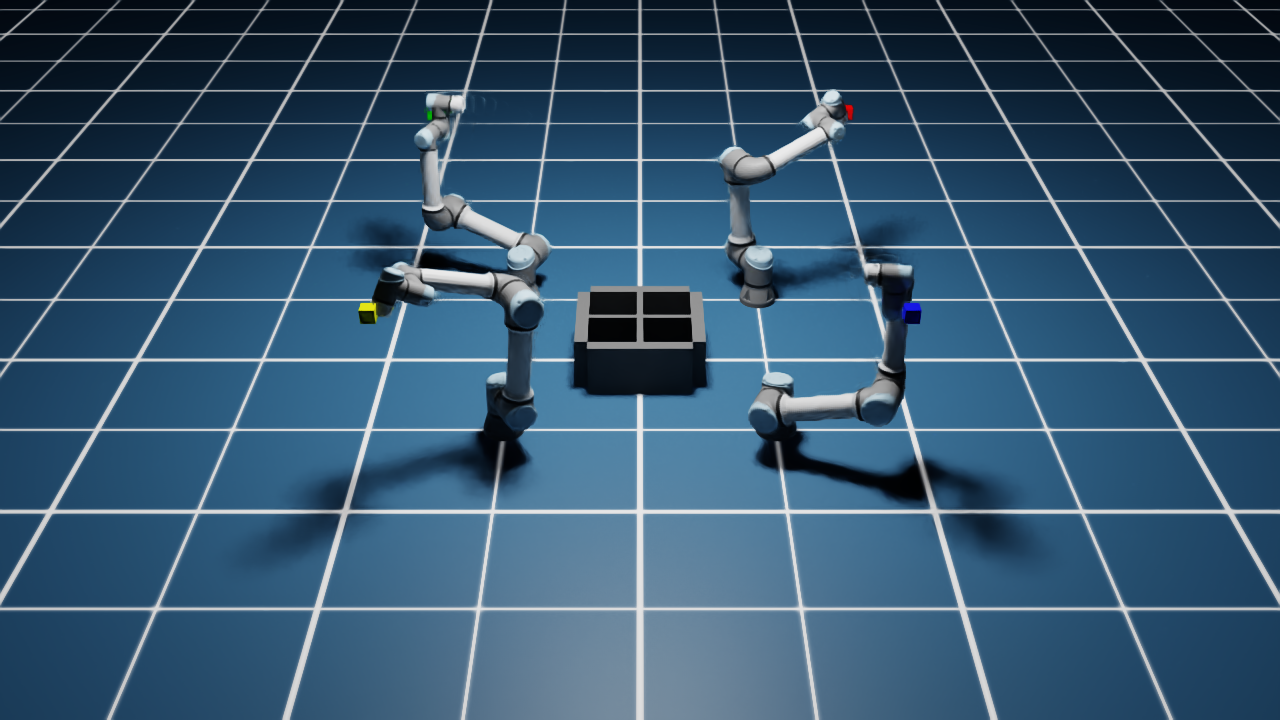}
        \vspace{-16pt}
        \caption*{A.\: Bin Loading}
    \end{subfigure}\hfill
    \begin{subfigure}{0.24\textwidth}
        \includegraphics[width=\linewidth, trim=350 270 400 120, clip]{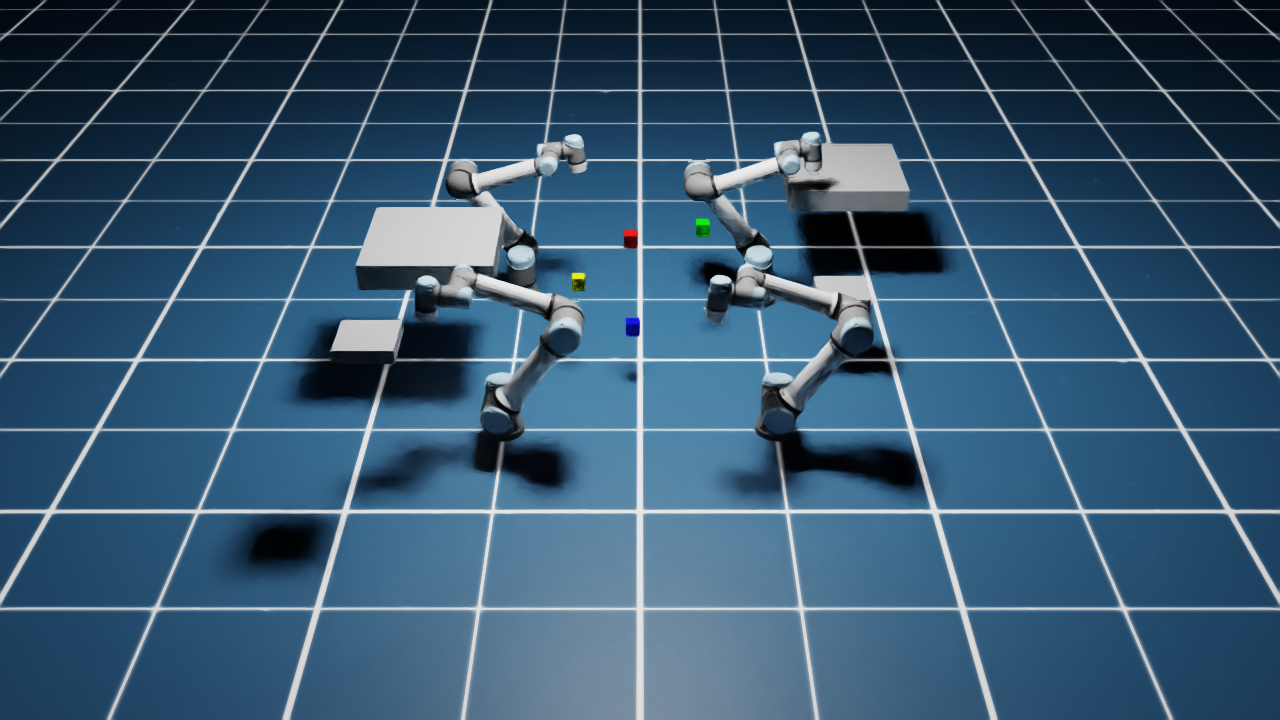}
        \vspace{-16pt}
        \caption*{B.\: Reaching-Hard}
    \end{subfigure}\hfill
    \begin{subfigure}{0.24\textwidth}
        \includegraphics[width=\linewidth, trim=400 280 380 130, clip]{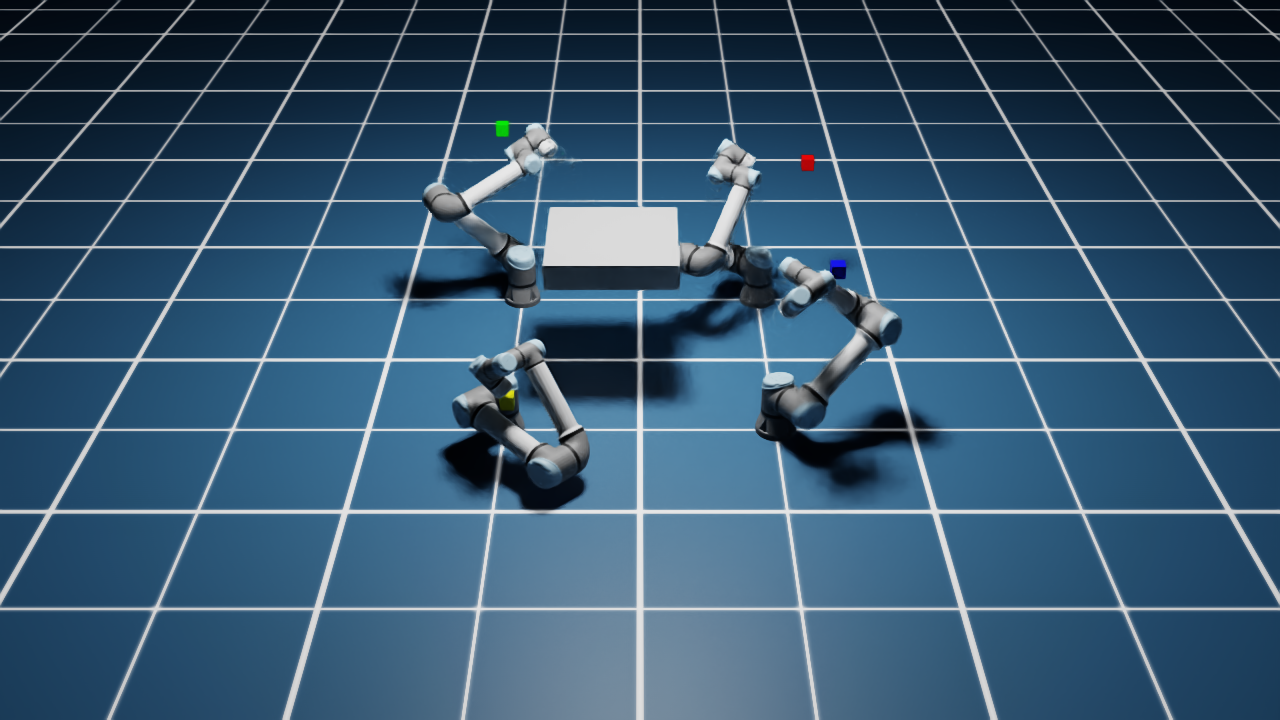}
        \vspace{-16pt}
        \caption*{C.\: Following}
    \end{subfigure}\hfill
    \begin{subfigure}{0.24\textwidth}
        \includegraphics[width=\linewidth, trim=5 330 35 100, clip]{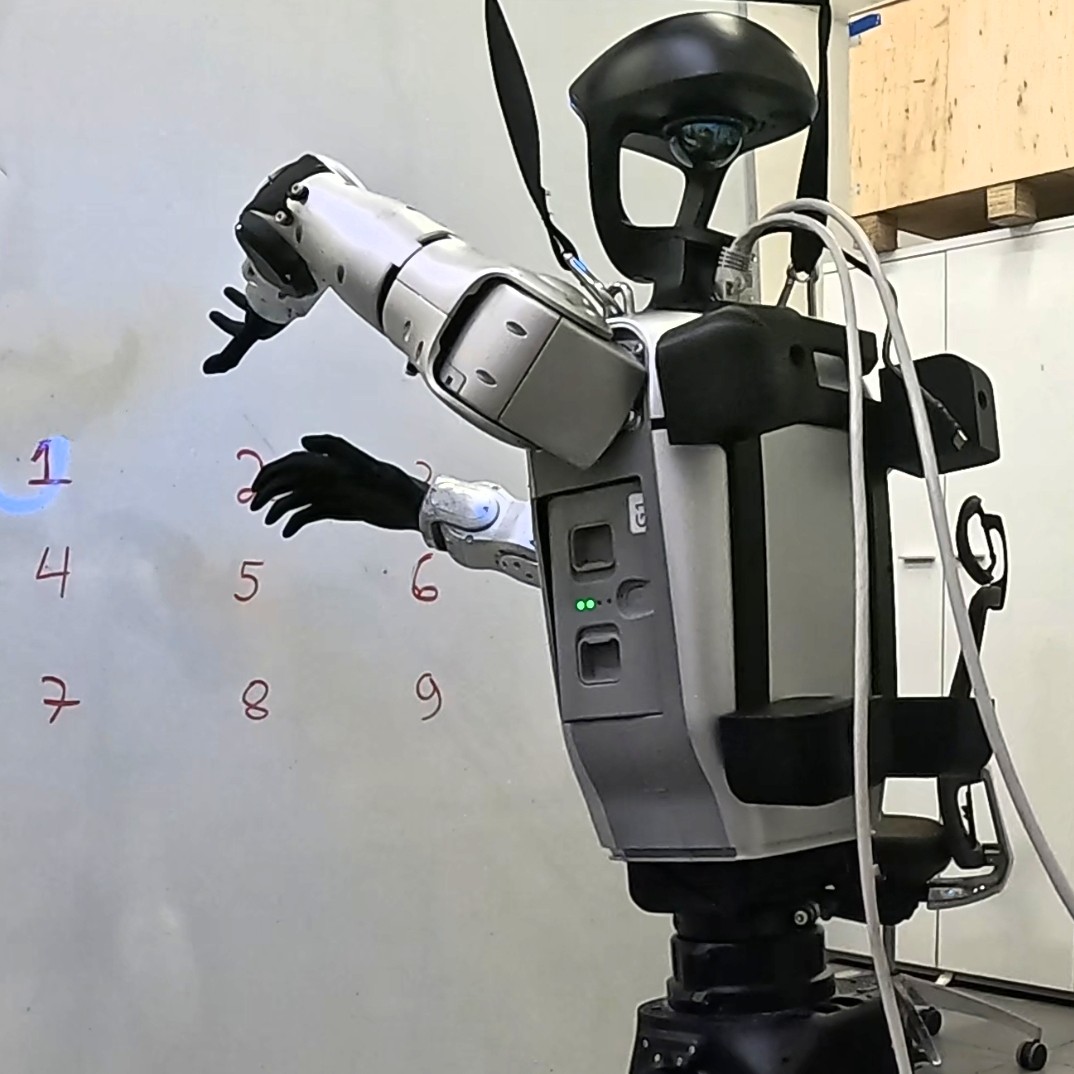}
        \vspace{-16pt}
        \caption*{D.\: Gridwall}
    \end{subfigure}
    
    \vspace{-7pt}
    \caption{\textbf{Task Setups.} A. \textit{Bin-Loading }initial state. Colored cubes: picking positions. B. \textit{Reaching-Hard} at initial state, level 5. Cuboids are dynamic and static obstacles. C. \textit{Following task}, level 2 (one obstacle). D. \textit{Gridwall}-humanoid controlled by MR-STORM. Right arm is prioritized (target=1). Left arm (target=6) avoids contact, letting right arm pass freely.}
    \label{fig:simulation_experiments}
    \vspace{-16pt}
\end{figure*}

\label{simulated_experiments:setup:scene_description}
In all trials, the arms were positioned at the corners of a $1\text{m}^2$ square (Fig.~\ref{fig:simulation_experiments}). Trials were halted after $T_{\max}=500$ simulation steps. Individual task details follow.

\label{simulated_experiments:setup:reaching}  

\textit{Reaching}: Goals reset upon arrival (5 cm tolerance) or after 1 s. There are static and dynamic obstacles with unknown trajectories, scaling in count with difficulty-level (Fig.~\ref{fig:simulation_experiments}B,C). \textit{Easy} samples goals locally (near arms) to minimize inter-arm interaction and focus on goal-change reactions. Conversely, \textit{Hard} samples goals centrally to maximize collision risk, compounding challenges through reduced path predictability and constrained free space.


 \label{simulated_experiments:setup:following}
\paragraph{Following} Each arm tracks a target, initially sampled locally, which then moves at constant velocity in an arbitrary direction; targets that drift out of range are reset. This task evaluates dynamic goal tracking (e.g., table cleaning) in multi-arm settings. As in reaching tasks, difficulty levels correspond to the number of obstacles present.

\label{simulated_experiments:setup:bin-loading}
\paragraph{Bin Loading (Fig.~\ref{fig:simulation_experiments}A)}
Arms transport objects from pickup spots (behind each robot) to four targets, 35 cm above a central bin. After picking, a valid target is randomly assigned based on the level rules. Difficulty ($l$) scales by concurrent access and target uniqueness: $l=1$ limits bin access to one arm; $l=2$ allows two arms with unique cells; $l=3$ permits all arms with unique cells; $l=4$ allows shared cells but only two concurrent arms; and $l=5$ removes all constraints, allowing all arms to target any cell simultaneously. This task evaluates performance under extreme workspace overlap and high-density collision risk.

\subsubsection{Metrics, Baselines, Reporting}

\label{simulated_experiments:metrics}
\paragraph{Metrics} 
Each environment is evaluated on safety and task performance. The collision score counts the simulation steps (out of $T_{\max}=500$) where any arm contacts an obstacle or another robot. Task scores vary by category: for \textit{reaching}, it is the cumulative goals reached by all arms; for \textit{bin-loading}, it is the total objects successfully deposited. For \textit{following}, we measure the mean positional tracking error:$${\mathrm{err_f}} = \frac{1}{T_{\max} \cdot A} \sum_{t=1}^{T_{\max}} \sum_{i=1}^{A} \| (ee_{p}^{(i)})_t - (g_{p}^{(i)})_t \|_2$$

\label{simulated_experiments:baselines}
\paragraph{Algorithms Tested}
We evaluate three MR-STORM variants against state-of-the-art (SOTA) GPU-accelerated motion planners. All methods observe current obstacle positions and are categorized into four families: MR-STORM, Centralized STORM, Decentralized STORM, and cuRobo.

\textbf{$\mathbf{M}$}: Our standard MR-STORM configuration. We use the default STORM library settings: $N=400$ rollouts and $K=5$ iterations. For all agents $i \in \{1, \dots, 4\}$, parameters are $H^{(i)}=40, w_{d}^{(i)}=5000, B^{(i)}=0.03$, and $\tau^{(i)}=3$.

\textbf{$\mathbf{M_{fast}}$:} This variant sets $K=1$ while keeping all other parameters identical to $\mathbf{M}$. Setting $K=1$ often suffices in STORM for high performance, and we seek to check whether this property holds for MR-STORM. The impact of iterations $K$ on control frequency is further detailed in \cite{Storm, Curobo}.

\textbf{$\mathbf{M}_{\tau}$:} Sets $\tau=0$ to isolate the impact of path communication by removing the prioritization scheme discussed in \ref{mr-storm:prioritization}. For clarity, this variant still shares plans but treats them with equal priority. All other parameters are identical to $\mathbf{M}$

\textbf{$\mathbf{C}, \mathbf{C^+}$:} As the SOTA for GPU-accelerated global motion planning, cuRobo \cite{Curobo} is applied here to the coupled 4-arm system. We evaluate two versions using the authors' recommended IK seed counts: $N_{\text{IKseeds}}=500$ ($\mathbf{C}$) and $1000$ ($\mathbf{C^+}$). For a fair comparison, a 5 cm position error tolerance was maintained across all algorithms.

\textbf{$\mathbf{S_c}, \mathbf{S_c^+}$:} Centralized STORM execution using a single controller for all arms. $\mathbf{S_c}$ ($N=400$) matches the per-arm rollout budget of $\mathbf{M}$, while $\mathbf{S_c^+}$ ($N=1600$) matches the total aggregate budget of the four arms. All other STORM parameters are identical to $\mathbf{M}$.

\textbf{$\mathbf{S_d}$:} Decentralized STORM. Each arm runs an independent controller treating other arms 
as static obstacles without prioritization or communication. They observe the current collision-spheres of others, but have no knowledge of their future plans. All other STORM parameters  identical to $\mathbf{M}$.

\label{simulated_experiments:reporting}
\paragraph{Reporting}
\label{simulated_experiments:reporting-statistical_method}
We report results using \emph{paired, per-environment differences} relative to our primary baseline $\mathbf{M}$ to eliminate nuisance variation caused by differing environment difficulties \cite{Demsar2006}. For each environment $e$ and algorithm $a$, we calculate the performance difference $\Delta T_{a,e}=T_{a,e}-T_{\mathbf{M},e}$ and collision difference $\Delta C_{a,e}=C_{a,e}-C_{\mathbf{M},e}$. These differences are aggregated across all 30 environments (5 difficulty levels $\times$ 6 environments), and we report the mean~$\pm$~SD. Following standard recommendations for multi-dataset comparisons \cite{Demsar2006}, we avoid data-dependent normalizations
to ensure statistical robustness.



\label{simulated_experiments:results}
\subsubsection{Results}

\begin{figure*}[t]
    \centering
    \includegraphics[width=\linewidth, trim=25 25 5 7, clip]{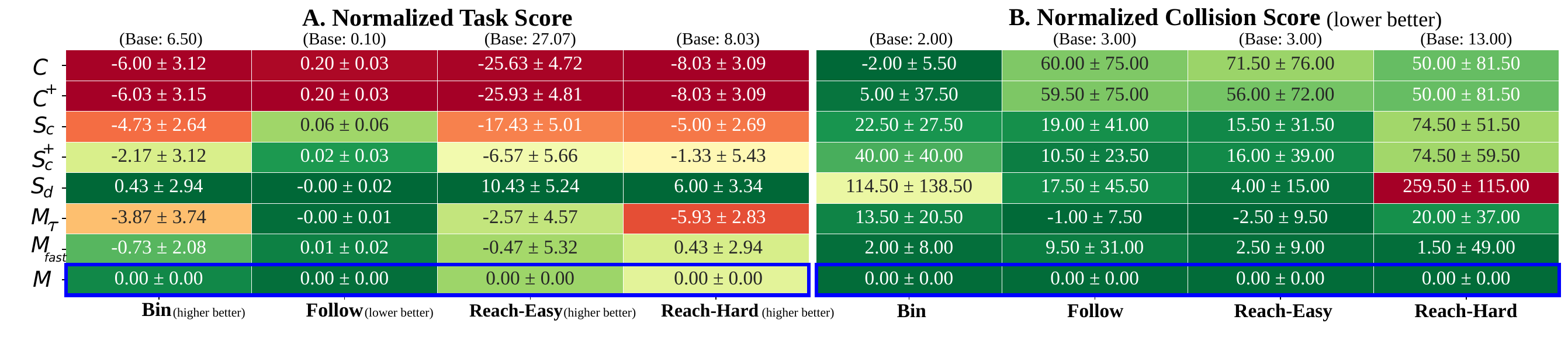}
    \vspace{-15pt}
   \caption{\textbf{Simulation Scores.} Heatmaps show normalized differences (A) \textbf{task scores} and (B) \textbf{collision scores} (mean $\pm$ SD, $N=30$) relative to baseline $\mathbf{M}$. Green/red gradients denote best/worst performance. Task scores are normalized per column; collision scores are normalized globally. \textbf{Baseline $\mathbf{M}$ (blue) includes raw mean values at the top of each column.}}
    \label{fig:tasks_and_safety}
\vspace{-4pt}
\end{figure*}

Figure~\ref{fig:tasks_and_safety} summarizes average task and collision scores across all tasks, with true baseline scores.  To analyze performance across difficulty levels, we compare the top-performing representative from each algorithmic family $\{\mathbf{C^+}, \mathbf{S_d}, \mathbf{S_c^+}\}$ against our baseline in Table \ref{tab:all_results_horizontal}.

\label{simulated_experiments:results}
\label{simulated_experiments:results:centralized_methods_struggle}

\paragraph{Conclusion 1: Centralized Methods Struggle in Dynamic Environments}
$\mathbf{C}$ and $\mathbf{C^{+}}$ exhibit the lowest performance. They respond to environmental changes at the same rate as other methods, but execute commands only after finding a global solution. This latency significantly degrades both task performance (Fig.~\ref{fig:tasks_and_safety}A) and safety (Fig.~\ref{fig:tasks_and_safety}B). While viable for simple scenarios (e.g., Table~\ref{tab:all_results_horizontal}, Bin, level 1), 
performance collapses as complexity increases: In bin-loading, significant workspace overlap prevents them from finding valid global solutions (Table~\ref{tab:all_results_horizontal}, Bin, levels 2-5). This issue is compounded in highly dynamic environments, where a decreasing plan-to-goal ratio leaves robots without actionable trajectories (Table~\ref{tab:all_results_horizontal}, Follow, levels~1-5). Doubling IK seeds ($\mathbf{C}$ to $\mathbf{C^{+}}$) does not resolve these fundamental limitations.

$\mathbf{S_c}$ and $\mathbf{S_c^+}$ outperform the $\mathbf{C}$ variants but still lag behind decentralized approaches. While $\mathbf{S_c^+}$ performs better than $\mathbf{S_c}$ due to its larger computational budget, both suffer from poor scalability as degrees of freedom (DoFs) increase. Notably, $\mathbf{S_c^+}$ and $\mathbf{S_d}$ utilize identical total computational resources. The superior performance of $\mathbf{S_d}$ suggests that a decentralized distribution of resources is far more efficient for discovering effective control strategies in multi-arm systems.

\label{simulated-experiments:results:overall-performance:decentralized}
\paragraph{Conclusion 2: Decentralization improves task performance but may compromise safety; plan communication mitigates this}
Decentralized methods ($\mathbf{M}$, $\mathbf{M}_{\text{fast}}$, $\mathbf{S_d}$) significantly outperform centralized ones in task performance (Fig.~\ref{fig:tasks_and_safety}A). But while $\mathbf{S_d}$ achieves high task scores, it is the least safe of all methods (Fig.~\ref{fig:tasks_and_safety}B). This safety gap occurs because independent arms react too late to inter-agent motions, causing catastrophic collisions — an effect particularly severe in high-density scenarios (see "MR-STORM vs Alternatives" on our website). For example, in \textit{reaching-hard}, $\mathbf{S_d}$ collision frequency is $\sim 20$ times higher than $\mathbf{M}$, 
with 272.5 average collision ratio for $\mathbf{S_d}$ vs.~only 13 for $\mathbf{M}$.

Across all algorithmic families, safety performance degrades significantly faster than in $\mathbf{M}$ as task difficulty and workspace density increase, particularly at levels 4 and 5. This trend is consistent across 11 out of 12 comparisons (Table \ref{tab:all_results_horizontal}) confirming that plan communication becomes increasingly critical in crowded environments. The only outlier is $\mathbf{C^+}$ in the bin-loading task, which maintains a stable collision rate; however, this is an artifact of the planner's failure to generate solutions, leaving the arms "frozen" and thus avoiding contact only by sacrificing all task progress.

\paragraph{Conclusion 3: Prioritization is essential for high-interaction tasks}
While $\mathbf{M}_{\tau}$ performs comparably to $\mathbf{M}$ and $\mathbf{M_{fast}}$ in low-interaction scenarios, its performance collapses in high-interaction environments—particularly in bin-loading and reaching-hard tasks. In the \textit{reaching-hard} task, $\mathbf{M}_{\tau}$ achieves only 2.1 goals on average, representing a 74\% performance drop compared to $\mathbf{M}$ (8.03) and $\mathbf{M_{fast}}$ (8.46) (Fig.~\ref{fig:tasks_and_safety}A). A similar 59\% decline is observed in the bin task. Notably, safety remains comparable across these three variants, suggesting that prioritization significantly boosts task efficiency without compromising the safety benefits provided by plan communication.

\paragraph{Conclusion 4: MR-STORM uniquely balances task success and safety}
MR-STORM (excluding $\mathbf{M}_{\tau}$) is the only family to consistently maintain high performance and safety across all metrics. This is visually striking in Fig.~\ref{fig:tasks_and_safety}, where the bottom two rows ($\mathbf{M_{fast}}$ and $\mathbf{M}$) are governed by green (high-performing) cells, while all other methods fail in at least one category. Notably, $\mathbf{M_{fast}}$ is highly efficient, nearly matching $\mathbf{M}$ at a significantly lower computational cost.

\begin{table*}[t]
\centering
\caption{\textbf{Per-level Simulation Scores} (normalized differences relative to $\mathbf{M}$). Each cell: mean $\pm$SD (Task / Collision).}
\label{tab:all_results_horizontal}
 \vspace{-4pt}
\scriptsize
\begin{tabular*}{\textwidth}{l l @{\extracolsep{\fill}} ccccc}
\toprule
\textbf{Task} & \textbf{Algo} & \textbf{Level 1} & \textbf{Level 2} & \textbf{Level 3} & \textbf{Level 4} & \textbf{Level 5} \\
\midrule
\multirow{3}{*}{\textbf{Bin-Load}} 
& $\mathbf{C^+}$ & -0.8$\pm$1.8 / 0.0$\pm$0.0 & -6.5$\pm$1.5 / 0.0$\pm$0.0 & -9.0$\pm$0.6 / -1.0$\pm$1.5 & -6.3$\pm$1.4 / 33.5$\pm$81.5 & -7.5$\pm$1.9 / -8.5$\pm$10.5 \\
& $\mathbf{S_c^+}$ & 1.0$\pm$2.7 / 4.0$\pm$10.0 & -1.3$\pm$2.0 / 46.5$\pm$58.5 & -5.0$\pm$3.3 / 80.5$\pm$20.0 & -2.3$\pm$2.4 / 24.0$\pm$28.5 & -3.2$\pm$2.1 / 45.5$\pm$25.0 \\
& $\mathbf{S_d}$ & 1.2$\pm$2.3 / 0.0$\pm$0.0 & 0.5$\pm$3.0 / 109.0$\pm$117.0 & 0.7$\pm$2.8 / 126.5$\pm$133.0 & -0.2$\pm$3.7 / 143.0$\pm$156.0 & 0.0$\pm$3.6 / 194.5$\pm$176.5 \\
\midrule
\multirow{3}{*}{\textbf{Reach-Easy}} 
& $\mathbf{C^+}$ & -28.0$\pm$5.2 / 0.0$\pm$0.0 & -26.5$\pm$2.9 / 22.5$\pm$55.5 & -26.8$\pm$2.7 / 22.5$\pm$55.5 & -25.7$\pm$4.4 / 85.0$\pm$61.5 & -22.7$\pm$7.3 / 150.0$\pm$48.0 \\
& $\mathbf{S_c^+}$ & -4.8$\pm$2.1 / 0.0$\pm$0.0 & -7.0$\pm$4.6 / 0.0$\pm$0.0 & -5.0$\pm$2.4 / 0.0$\pm$0.0 & -3.5$\pm$6.0 / -1.0$\pm$2.0 & -12.5$\pm$7.6 / 81.5$\pm$49.5 \\
& $\mathbf{S_d}$ & 11.2$\pm$5.5 / 0.0$\pm$0.0 & 11.8$\pm$5.9 / 0.0$\pm$0.0 & 11.0$\pm$5.6 / 0.0$\pm$0.0 & 10.5$\pm$3.7 / 4.0$\pm$12.5 & 7.7$\pm$6.1 / 17.0$\pm$30.5 \\
\midrule
\multirow{3}{*}{\textbf{Reach-Hard}} 
& $\mathbf{C^+}$ & -10.3$\pm$1.6 / -18.5$\pm$33.0 & -8.5$\pm$3.1 / -3.0$\pm$5.0 & -7.7$\pm$4.3 / -9.0$\pm$15.0 & -8.0$\pm$2.7 / 135.5$\pm$4.5 & -5.7$\pm$1.9 / 144.5$\pm$68.5 \\
& $\mathbf{S_c^+}$ & 1.0$\pm$6.3 / 78.5$\pm$68.0 & -2.3$\pm$6.9 / 74.5$\pm$27.0 & -1.7$\pm$5.6 / 72.5$\pm$38.5 & -3.2$\pm$2.5 / 88.0$\pm$53.5 & -0.5$\pm$5.7 / 58.0$\pm$102.0 \\
& $\mathbf{S_d}$ & 6.2$\pm$3.8 / 318.0$\pm$102.5 & 6.3$\pm$5.5 / 296.0$\pm$106.0 & 6.5$\pm$3.7 / 279.0$\pm$73.0 & 5.0$\pm$1.5 / 224.5$\pm$95.5 & 6.0$\pm$1.4 / 179.0$\pm$157.5 \\
\midrule
\multirow{3}{*}{\textbf{Follow}} 
& $\mathbf{C^+}$ & 0.2$\pm$0.0 / 0.0$\pm$0.0 & 0.2$\pm$0.0 / 0.0$\pm$0.0 & 0.2$\pm$0.0 / 0.0$\pm$0.0 & 0.2$\pm$0.0 / 137.0$\pm$1.5 & 0.2$\pm$0.0 / 161.0$\pm$19.5 \\
& $\mathbf{S_c^+}$ & 0.0$\pm$0.0 / 0.0$\pm$0.0 & 0.0$\pm$0.0 / 0.0$\pm$0.0 & 0.0$\pm$0.0 / 0.0$\pm$0.0 & 0.0$\pm$0.0 / 0.5$\pm$1.0 & 0.0$\pm$0.1 / 52.5$\pm$23.5 \\
& $\mathbf{S_d}$ & -0.0$\pm$0.0 / 0.0$\pm$0.0 & -0.0$\pm$0.0 / 0.0$\pm$0.0 & -0.0$\pm$0.0 / 0.0$\pm$0.0 & -0.0$\pm$0.0 / 0.0$\pm$0.0 & 0.0$\pm$0.0 / 86.5$\pm$69.5 \\
\bottomrule
\end{tabular*}
\end{table*}

\label{real_experiments}
\subsection{Real-Arms Experiments}
\label{real_experiments:setup}
\subsubsection{Setup}
We evaluate MR-STORM on a Unitree G1™ ($d=29$) humanoid robot by controlling both 7-DoF arms,
fixing the remaining 15 joints. Due to simulated safety risks observed in alternative baselines, only MR-STORM was deployed.
Because the torso and legs are fixed, we model the rest of the body parts as static environment obstacles to prevent inter-limb and self-collisions. 

Running on the same PC used for the simulated experiments, commands for each arm were generated independently via two separate instances of Alg.~\ref{alg:MRSTORM} and transmitted to the robot over an Ethernet cable. 
Prior to hardware deployment, we manually fine-tuned the parameters of MR-STORM for each arm, starting from the baseline UR5e embodiment configuration used in simulation~(\ref{simulated_experiments:setup:robot_spec}). To maintain consistency with the simulated experiments, we used a similar planning budget of $N^{(i)}=400$ and $H^{(i)}=30$, while the fine-tuned parameters were set to $B=0.03$, $\tau^{(i)}=10$, and $w_d^{(i)}=300$ for each arm $i$. Since the physical robot operates at a control frequency of $30$\,Hz per arm, we adjusted the simulated time between steps in both simulator and MPPI model (as part of Alg.~\ref{alg:MRSTORM}) to $\frac{1}{30}$s accordingly.

\textbf{Task:} We implement a ``Gridwall'' pseudo-pick-and-place task (Fig.~\ref{fig:simulation_experiments}.D). The arms must navigate between lateral ``picking'' stations near object heaps and a central $3\times3$ target gridwall ($20\times30$\,cm) containing 9 placement spots numbered 1–9. Target spots are randomly sampled, forcing frequent arm crossovers within a highly conflictual workspace. We conducted 10 trials of $100$\,s each 
with a $5$\,cm spot-reaching tolerance, matching the simulated evaluation~(\ref{simulated_experiments:setup:reaching}). 
Although the fixed end-effectors lack finger joints for actual grasping, the method generalizes to full manipulation by incorporating object payloads into the robot's collision geometry, as demonstrated in STORM.

\label{real_experiments:metrics}
\textbf{Metrics:} To measure performance, we evaluate the total number of pseudo-placed items by both arms (similar to \ref{simulated_experiments:setup:bin-loading}), as well as the number of collisions during each trial. We report the $\min$, $\max$, and $\text{mean} \pm \text{SD}$ of the number of successfully placed items.

\subsubsection{Results}

Table~\ref{tab:real_world_results} summarizes the results. MR-STORM performed consistently across seeds with balanced arm usage. Safety was robust; no high-impact collisions occurred over 1000\,s. The only contact was a single minimal fingertip touch (seed 3), likely due to simplified collision spheres modeling. Videos are available on our site.

\begin{table}[h]
\centering
\caption{Real-World Arm Experiments: MR-STORM performance across $N = 10$ seeds (100 seconds per seed).}
\label{tab:real_world_results}
\begin{tabular}{lcccc}
\toprule
\textbf{Metric} & \textbf{Mean ($\pm$ SD)} & \textbf{Min} & \textbf{Max} \\ \midrule
Placements-Left Arm & $7.2 \pm 2.0$ & 5 & 11  \\
Placements-Right Arm& $8.3 \pm 1.5$ & 6 & 11  \\
Placements-Two Arms & $15.5 \pm 3.1$ & 11 & 21 \\
Collisions & $0.1 \pm 0.3$ & 0 & 1  \\ \bottomrule
\end{tabular}
\vspace{-5pt}
\end{table}

\section{Summary}
We presented MR-STORM, extending the STORM algorithm to multi-arm systems via distance-based prioritization and a modified cost function. Across diverse simulation environments and real-world hardware, MR-STORM consistently outperforms benchmarks with minimal tuning required for platform transfer, providing the best balance of task performance and safety.

\section{Limitations and Future Work}
\label{Section: limitations}
MR-STORM inherits several limitations from the base STORM framework: it is a local MPC-based method lacking global optimality, utilizes soft constraints that do not strictly guarantee safety and joint limits, models environmental obstacles as static in the MPPI dynamics model, and employs coarse sphere-based occupancy representations. Future work should explore rigorous safety enforcement, for example, by pruning unsafe trajectories within the MPPI loop.
Our work assumes short communication latency; future iterations must address the impact of network delays and packet loss on the algorithm's reliability.

\bibliographystyle{IEEEtran}
\bibliography{IEEEabrv,mybibfile}





\end{document}